\documentclass{sig-alternate-05-2015}
\usepackage{times}
\usepackage{url}
\usepackage{latexsym}

\usepackage{times}
\usepackage{url}
\usepackage{latexsym}
\usepackage{graphicx}
\usepackage{url}
\usepackage{algorithm}
\usepackage{algorithmic}
\usepackage{subcaption}
\usepackage{multirow}
\usepackage{comment}
\usepackage{amssymb}
\usepackage{amsmath}
\usepackage{CJKutf8}
\begin{document}
%
\CopyrightYear{2016} 
\setcopyright{acmcopyright}
\conferenceinfo{ICMR'16,}{June 06-09, 2016, New York, NY, USA}
\isbn{978-1-4503-4359-6/16/06}\acmPrice{\$15.00}
\doi{http://dx.doi.org/10.1145/2911996.2912016}

\title{Multilingual Visual Sentiment Concept Matching}

\numberofauthors{5}

\author{
\alignauthor \hspace{-10mm} Nikolaos Pappas\\
       \affaddr{\hspace{-10mm}Idiap Research Institute}\\
       \affaddr{\hspace{-10mm}Martigny, Switzerland}\\
       \email{\hspace{-10mm}npappas@idiap.ch}
\alignauthor\hspace{-30mm} Miriam Redi\\
       \affaddr{\hspace{-30mm} Yahoo Inc.}\\
       \affaddr{\hspace{-30mm} London, United Kingdom}\\
       \email{\hspace{-30mm} redi@yahoo-inc.com}
\alignauthor \hspace{-45mm} Mercan Topkara\\
       \affaddr{\hspace{-45mm} JW Player}\\
       \affaddr{\hspace{-45mm} New York, NY, USA}\\
       \email{\hspace{-45mm} mercan@jwplayer.com}
\alignauthor \hspace{-55mm} Brendan Jou\\
       \affaddr{\hspace{-55mm} Columbia University}\\
       \affaddr{\hspace{-55mm} New York, NY, USA}\\
       \email{\hspace{-55mm} bjou@ee.columbia.edu}   \and
\alignauthor Hongyi Liu\\
       \affaddr{Columbia University}\\
       \affaddr{New York, NY, USA}\\
       \email{hongyi.liu@columbia.edu}
\alignauthor Tao Chen\\
       \affaddr{Columbia University}\\
       \affaddr{New York, NY, USA}\\
       \email{taochen@ee.columbia.edu}
\alignauthor Shih-Fu Chang\\
       \affaddr{Columbia University}\\
       \affaddr{New York, NY, USA}\\
       \email{sfchang@ee.columbia.edu}
}

\maketitle

\begin{abstract}
The impact of culture in visual emotion perception has recently captured the attention of multimedia research. In this study, we provide powerful computational linguistics tools to explore, retrieve and browse a dataset of 16K multilingual affective visual concepts and 7.3M Flickr images. First, we design an effective crowdsourcing experiment to collect human judgements of sentiment connected to the visual concepts. We then use word embeddings to represent these concepts in a low dimensional vector space, allowing us to expand the meaning around concepts, and thus enabling  insight about commonalities and differences among different languages. We compare a variety of concept representations through a novel evaluation task based on the notion of \emph{visual semantic relatedness}. Based on these representations, we design clustering schemes to group multilingual visual concepts, and evaluate them with novel metrics based on the crowdsourced sentiment annotations as well as visual semantic relatedness. The proposed clustering framework enables us to analyze the full multilingual dataset in-depth and also show an application on a facial data subset, exploring cultural insights of portrait-related affective visual concepts.
\end{abstract}


\section{Introduction}
Everyday, billions of users from all around the world share their visual memories in online photo sharing platforms. Web users speak hundreds of different languages, come from different countries and backgrounds.  Such multicultural diversity also results in users representing the visual world in very different ways. For instance, \cite{Jou2015} showed that Flickr users with different cultural backgrounds use different concepts to describe visual emotions. But how can we build tools to analyze and retrieve the multimedia data generated by such a diverse population?

Multimedia retrieval in a multilingual and multicultural environment cannot be independent of the spoken language used by users to describe their visual content.
\begin{figure}[t]
\centering
\vspace{-1mm}
\includegraphics[width=3.3in]{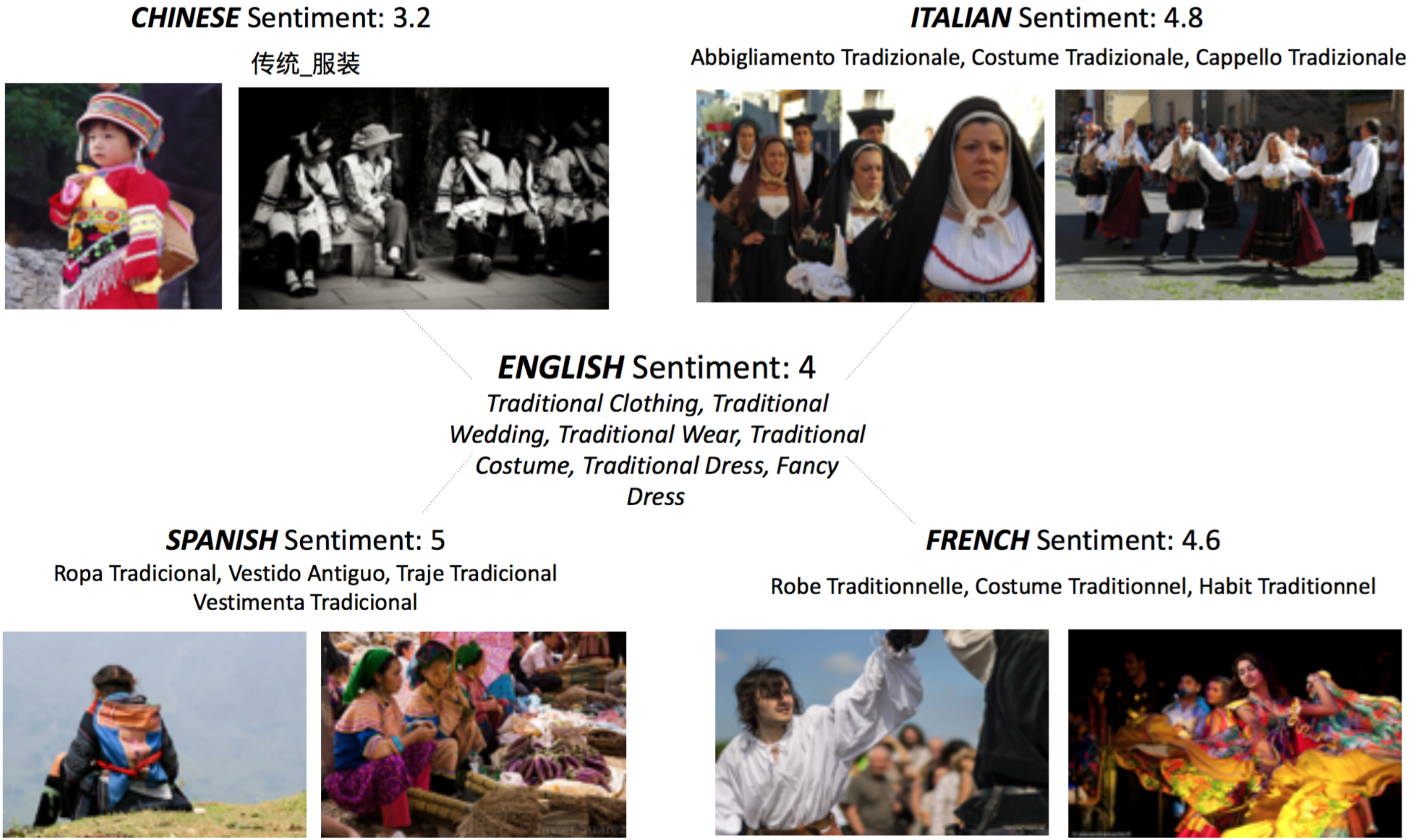}
\vspace{-4mm}
\caption{Example images from four languages from the same ``traditional clothing'' cluster. Even though all images are tagged with semantically similar concepts, each culture interprets such concepts with different visual patterns and sentimental values.}
\vspace{-6mm}
\label{fig:example}
\end{figure}
For example, among the vast content of photo sharing platforms such as in Flickr, it is easy to find pictures of traditional costumes from all around the world. However, a basic keyword search, e.g.~\emph{traditional costumes}, would not return multiculturally rich results: the returned content would very likely come from Western countries, especially from countries where English is the primary language. The problem we tackle is to analyze, index and develop a deeper understanding of multicultural content in the context of a large social photo sharing platform. A purely image-based analysis would not provide a complete understanding as it clusters visually-similar images together, missing the differences between cultures, e.g.~how an \emph{old house} or \emph{good food} look in each culture. We mitigate the problems of pure image-based analysis with the aid of analytical language tools.

In this paper, we focus on one dimension which characterizes users' cultural background, namely their spoken language, and we build the first complete framework for analyzing, exploring, and retrieving multilingual emotion-biased visual concepts. This allows to retrieve examples of concepts such as \emph{traditional costumes} from visual collections of different languages (see Fig.~\ref{fig:example}). Our goal is to study the lexical differences of sentiment-biased visual concepts across multiple languages from the MVSO dataset \cite{Jou2015}. Complementarily, we also seek to investigate the visual differences for images related to similar visual concepts across languages. 
To achieve this, it is essential to match lexical expressions of concepts from one language to another. One way to achieve this is through \emph{exact matching}, arising from the translation of all languages to a single one, e.g.~English.
However, given that lexical choices for the same concepts vary across languages, the exact matching of multilingual concepts has a small coverage across languages. To overcome this sparsity issue, we propose an \emph{approximate matching} approach. We represent multilingual concepts in a common semantic embedding based on pre-trained word embeddings. This allows us to compute the semantic proximity or distance between visual concepts and cluster concepts from multiple languages. Furthermore, the approximate matching enables a better connectivity between visual concepts of different languages. This allows discovery of multilingual clusters of visual concepts, whereas exact matching clusters are mostly dominated by a single language.

The contributions of our work can be summarized as follows:
\\(1) We design a crowdsourcing experiment to annotate the sentiment score of visual concepts from 11 languages in MVSO, and make the resulting data publicly available.
\\(2) We evaluate and compare a variety of unsupervised distributed word and concept representations. To do so, we define a novel evaluation metric called \emph{visual semantic relatedness}, defined as the cosine distance of the multilingual concept co-occurrence on real-world images from MVSO.
By analyzing sentiment values across languages and their translations, we prove the need of effective multilingual sentiment analysis tools.
\\(3) We design new tools to evaluate sentiment and semantic consistency on various visual concept clustering techniques 
that allow us to organize MVSO in an hierarchical manner by obtaining rich multilingual clusters of affective concepts.
\\(4) We demonstrate the applicability of the proposed approaches with qualitative analysis and a novel case study on analyzing portrait images in MVSO.
We find that Eastern and Western languages tend to attach different visual concepts to portrait images, but that many languages similarly attach positive concepts to face pictures.

\section{Related Work}

Research on distributed word representations \cite{turian10,collobert11,word2vec,glove} have recently extended to multiple languages either by using bilingual word alignments or parallel corpora to transfer linguistic information from multiple languages. For instance, \cite{rfou13} proposed to learn distributed representations of words across languages by using a multilingual corpus from Wikipedia. \cite{klementiev12,zou13} proposed to learn bilingual embeddings in the context of neural language models utilizing multilingual word alignments. \cite{hermann14} proposed to learn joint-space embeddings across multiple languages without relying on word alignments. Similarly, \cite{chandar14} proposed auto-encoder-based methods to learn multilingual word embeddings. A limitation when dealing with many languages is the scarcity of data for all pairs. In this study, we use a pivot language to align the multiple languages using machine translation. 

To our knowledge, the visual context of sentiment concepts for evaluating multilingual embeddings have not been considered before. However, studies on multimodal distributional semantics have combined visual and textual features to learn more informed word embeddings and have used the notion of semantics \cite{hill14,bruni14} and visual similarity to evaluate word embeddings \cite{silberer14,lazaridou15}. Furthermore, there are studies which have combined language and vision for image caption generation and retrieval \cite{karpathy14,kiros14,socher14,mao14} based on multimodal neural language models. We argue that our evaluation metric is a rich resource for learning more informed multimodal embeddings which can benefit these systems. Perhaps the most related study to ours is in \cite{kottur15} which aimed to learn visually grounded word embeddings to capture visual notions of semantic relatedness using abstract visual scenes. Our differences are that we focus on sentiment concepts and we define visual semantic relatedness based on real-world images which are annotated by community users of Flickr instead of abstract scenes.

\section{Visual Sentiment of Concepts} \label{sec:crowdflower}

In this work, we make use of the MVSO dataset \cite{Jou2015}.
One drawback of MVSO is that the sentiment scores assigned to affective visual concepts were automatically inferred through sentiment analysis tools.
Although such tools have achieved impressive performances in the recent years, they are typically based on text modalities alone. To counter this, we designed a crowdsourcing experiment with CrowdFlower\footnote{\url{http://www.crowdflower.com}} to annotate the sentiment of the multilingual adjective-noun pairs (ANPs) in MVSO. We considered 11 out of 12 languages in MVSO, leaving out Persian due to the limited number of ANPs. We constructed separate tasks for each language, using all ANPs in MVSO for that language. 
\vspace{3pt}
\\\textbf{Task Interface.} We asked crowdsourcing workers to evaluate the sentiment value of ANPs on a scale from 1 to 5. We provided annotators with intuitive instructions, along with examples ANPs with different sentiment values. Each task showed five ANPs from a given language along with Flickr images associated to those ANPs. Annotators rated the sentiment expressed by each ANP, choosing between ``very negative,'' ``slightly negative,'' ``neutral,'' ``slightly positive'' or ``very positive''.
\vspace{3pt}
\\\textbf{Task Setup.}
The sentiment of each ANP was judged by five or more independent workers.
Similar to the MVSO setup, we required that workers were both native speakers of the task's language and highly ranked on the platform.
\begin{figure}[t]
\centering
\includegraphics[width=\linewidth]{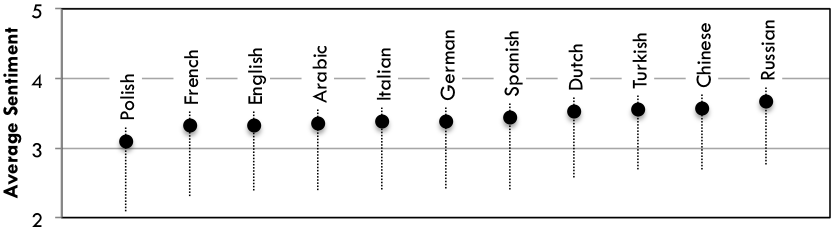}
\vspace{-6mm}
\caption{Crowdsourced visual sentiment per language in MVSO.} 
\label{fig:crowdflower:sentiment}
\vspace{-5mm}
\end{figure}
We also developed a subset of test questions with an expert-labeled gold standard: to access a crowdsourcing task, workers needed to correctly answer 7 of 10 test questions. Their performance was also monitored throughout the task by randomly inserting a test question in each task.
\vspace{3pt}
\\\textbf{Crowdsourcing Results.}
To assess the quality of the collected annotations, we computed the level of agreement between contributors. And although sentiment assessment is intrinsically a subjective task, we found an average agreement of $\sim$68\%. We found an average correlation of 0.54 between crowdsourced sentiment scores and the automatically assigned sentiment scores in \cite{Jou2015}. Although this value is reasonably high, it still shows that the two sets of scores do not completely overlap. A high-level summary of the average sentiment collected per language is shown in Figure \ref{fig:crowdflower:sentiment}. Unlike the findings of previous work using automatic sentiment scores \cite{Jou2015}, we find that visual concepts from Chinese and Russian users carry more positive sentiment on the average when compared to concepts shared by speakers of other languages.

\begin{figure*}[htp]
  \vspace{-8mm}
  \centering
  \hspace{-2mm}\includegraphics[scale=0.34]{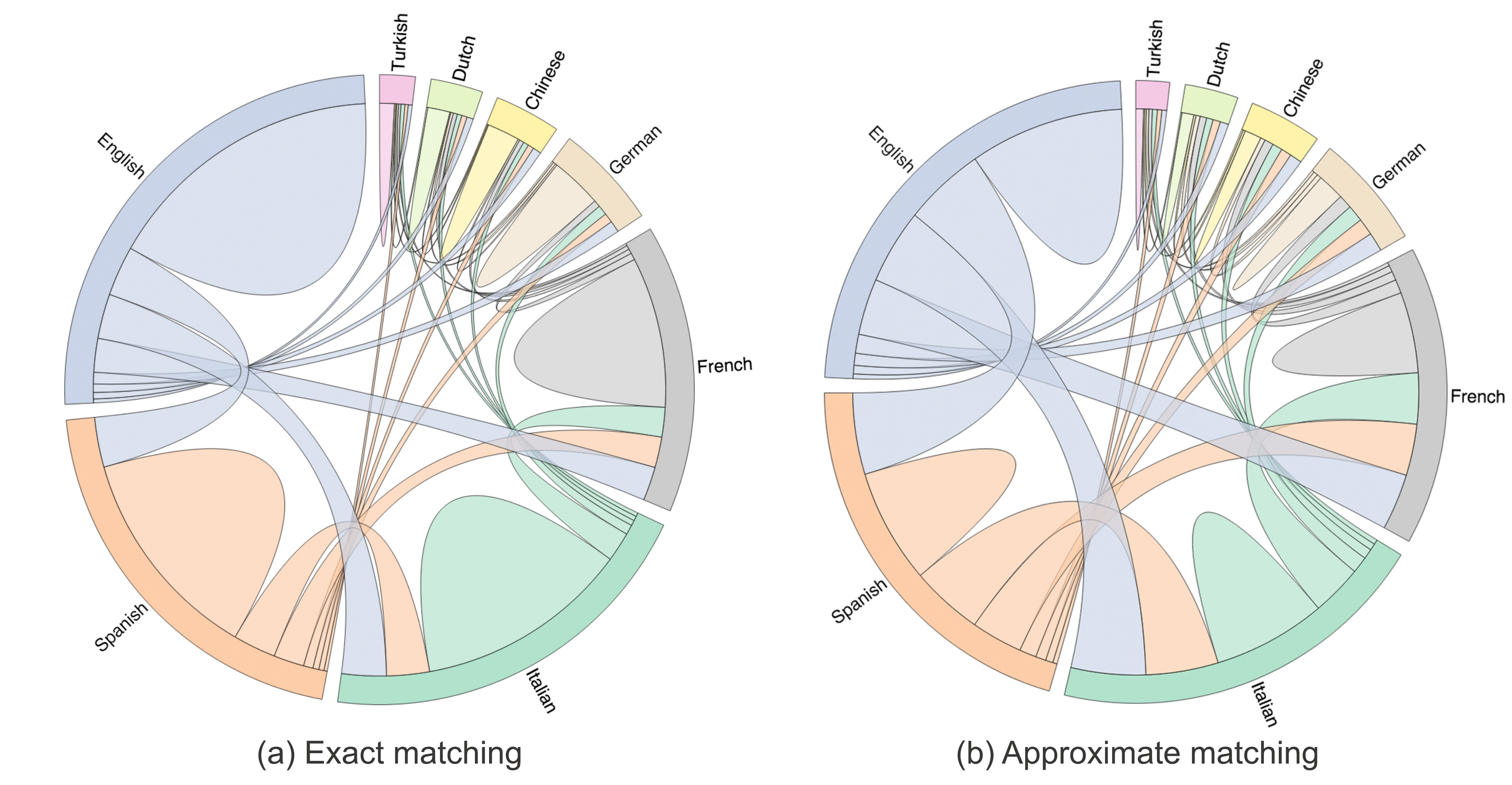} 
  \vspace{-4mm}
  \caption{Clustering connectivity across top-8 most popular languages in MVSO measured by the number of concepts in the same cluster of a given language with other languages represented in a chord diagram.   On the left (a), the clusters based on exact matching are mostly dominated by a single language, while on the right (b), based on approximate matching, the clusters are enriched with multiple languages.  }
  \vspace{-3mm}
  \label{matching}
\end{figure*} 

\section{Exact Concept Matching}
To match visual sentiment concepts (ANPs) across languages we translated them from each language $L = \{l_ij\ |\ i=1 \ldots m, j=1 \ldots n_{i}\}$
to the concepts of a pivot language $C = \{c_{i}\ |\ i=1 \ldots N\}$ using the Google Translate API\footnote{\url{https://cloud.google.com/translate}}. We selected English as the pivot language because it has the most complete translation resources (parallel corpora) for each of the other languages due to its popularity in relevant studies. For instance, the concepts \emph{chien heureux} (French), \emph{perro feliz} (Spanish) and \emph{gl\"{u}cklicher hund} (German) are translated to the English concept \emph{happy dog}. Rightly so, one would expect that the visual sentiment concepts in the pivot language might have shifted in terms of sentiment and meaning as a result of the translation process. And so, we examine and analyze the effects of translation to the sentiment and meaning of the multilingual concepts as well as the matching coverage across languages.

\subsection{Sentiment and Meaning Shift}

To quantitatively examine the effect of translation to the sentiment of concepts, we used the crowdsourced sentiment values and count the number of concepts for which the sentiment is shifted after translation in English. The higher this number for a given language, the higher the specificitiy of the visual sentiment for that language. To avoid counting sentiment shifts caused by small sentiment values we define a sentiment threshold $t$ below which we do not consider sign changes, as follows: $|\textrm{sent}(c_{i})| > t$. Table \ref{sent_shift} displays the percentage of concepts with shifted sign due to translation. The percentages are on average about 33\% for $t=0$. The highest percentage of sign shift is 60\% from Arabic and the lowest percentage is 18.6\% for Dutch. Moreover, the percentage of concepts with shifted sign decreases for most languages as we increase the absolute sentiment value threshold $t$ from 0 to 0.3.  This result is particularly interesting since it suggests that visual sentiment understanding can be enriched by considering the language dimension.

The translation can affect also the meaning of the original concept in the pivot language. For instance, a concept in the original language which has intricate 
compound words (adjective and noun) could be translated to simpler compound words. This might be due to the lack of expressivity of the pivot language, or to compound words with shifted meaning, because of translation mistake, language idioms, or lack of large enough context. For example, \begin{CJK}{UTF8}{bkai}民主法治\end{CJK} (Chinese) is translated to \emph{democracy and the rule of law} in English, 
while \emph{passo grande} (Italian) is translated to \emph{plunge} and \emph{marode sch\"{o}nheit} (German) is translated in to \emph{ramshackle beauty}. 
Examining the extent of this effect
\begin{table}[htp]
\footnotesize
  \centering
  \begin{tabular}{| l | l | l | l | l|}
  \hline
  \textbf{Language} & $t=0.0$ & $t=0.1$ & $t=0.2$ & $t=0.3$  \\ \hline      
        Italian & 28.9 (6.0) & 16.7 (3.3) & 11.4 (2.4) & 7.3 (2.2) \\
        French  & 36.2 (8.1) & 23.6 (5.3) & 16.8 (3.8) & 9.7 (3.3) \\
        Chinese & 24.4 (6.3) & 11.8 (5.5) & 5.5  (1.4) & 3.1 (0.8) \\
        German  & 27.1 (6.2) & 15.5 (3.5) & 8.3  (1.9) & 7.7 (1.8) \\
        Dutch   & 18.6 (5.4) & 8.2  (2.4) & 6.2  (1.8) & 3.1 (0.9) \\
        Russian & 25.6 (8.3) & 20.5 (6.6) & 5.1  (1.7) & 2.6 (0.8) \\
        Turkish & 33.3 (8.2) & 22.2 (5.5) & 7.4  (1.8) & 3.7 (0.9)  \\
        Arabic  & 60.0 (21.4) & 40.0 (14.3)& 10.0 (3.6) & 10.0 (3.6) \\ \hline
  \end{tabular}
  \vspace{-2mm}
  \caption{Percentage of concepts with sentiment sign shift after exact match and over all ANPs in a given language (in parenthesis).   }
  \label{sent_shift}
  \vspace{-5mm}
\end{table}
  quantitatively is very costly because it requires language experts from all languages at hand. Sentiment and meaning shift effects are very interesting open problems, however, in this study,
we focus specifically on how to increase the matching coverage among languages 
rather than what is the optimal modeling of sentiment and meaning, hence we leave them as future work.

\subsection{Matching Coverage}
The matching coverage is an essential property for multilingual concept matching and clustering. We examine this property by performing $k$-Means clustering of multilingual concepts, and counting the number of concepts between two languages that belong to the same cluster. This reveals the  connectivity of language clusters based on exact matching, as shown in Figure \ref{matching} (a) for the top-8 most popular languages in MVSO. From the connection stripes which represent the number of concepts between two languages, we can observe that, when using exact matching, concept clusters are 
dominated by single languages. For instance, in all the languages there is a connection stripe which connects back to the same language: this means that many clusters contain monolingual concepts. Another example which highlights the disadvantage of exact matching is that 
out of all the German translations (781), the ones matched with  Dutch concepts (39) were more numerous than the ones matched with  Chinese concepts (23). This was striking given that there were less (340) translations from Dutch than from Chinese (472). This demonstrates that the matching of concepts among languages is generally very sparse and does not depend necessarily on the number of translated concepts; this hinders our ability to compare concepts across languages in a unified manner. 
 
Moreover, we want be able to know the relation among concepts from original languages where we cannot have a direct translation. To address the limitations of exact matching coverage of concepts, we propose below to expand the exact match translation and approximate translations based on compound word semantics.

\section{Approximate Concept Matching}
We saw that matching sentiment-biased visual concepts in different languages with exact matching has several limitations, in terms of translation correctness and concept coverage. To overcome these limitations, we relax the exact condition for matching multilingual concepts, and instead we \emph{approximately} match concepts  based on their semantic meaning. Intuitively, in order to match concepts from different languages, we need a proximity (or distance) measure reflecting how `close' or similar concepts are in the semantic distance space. This enables to achieve our main goal: comparing visual concepts cross-lingually, and cluster them in to multilingual groups.  When using, approximate matching the clustering connectivity between languages is greatly enriched, as shown in Figure \ref{matching} (b): connection stripes are more evenly distributed for all languages. To learn such representations of meaning we make use of the recent advances in distributional lexical semantics \cite{word2vec,glove,eval15,goldberg15} utilizing the skip-gram model provided by word2vec toolkit\footnote{\url{https://code.google.com/p/word2vec}} trained on large text corpora.

\subsection{Distributed Word Representations}
\label{sec:embed}
To represent words in a semantic space we use unsupervised word embeddings based on the skip-gram model via word2vec. Essentially, the skip-gram model aims to learn vector representations for words by predicting the context of a word in a large corpus. The context is defined as a window of $w$ words before and $w$ after the current word. We consider the following corpora in English on which the skip-gram model is trained:

\begin{enumerate}
  \item \textbf{Google News}: A corpus of news which contains 100 billion tokens and 3,000,000 unique words which have at least 5 occurrences from \cite{mikolov13}. News are typically used to describe real-world events which contains accurate word meanings, however it has indirect relevance to visual content.
  \item \textbf{Wikipedia}: A corpus of Wikipedia articles which contains 1.74 billion tokens and 693,056 unique words which have at least 10 occurrences. The pre-processed text of this corpus was obtained from \cite{lebretL13}. 
  Wikipedia articles are more thorough descriptions of real-world events, entities, objects, and concepts. Similar to Google news, the visual content is indirectly connected to the word usage. 
  \item \textbf{Wikipedia + Reuters + Wall Street Journal}: A mixture corpus of Wikipedia articles, Wall Street Journal (WSJ) and Reuters news which contains 1.96 billion tokens and 960,494 unique words which have at least 10 occurrences. The pre-processed text of this corpus was obtained from \cite{lebretL13}. This combination of news articles and Wikipedia articles captures a balance between these two different types of word usage.
  \item \textbf{Flickr 100M}: A corpus of image metadata which contains 0.75 billion tokens and 693,056 unique words (with frequency higher than 10) available from Yahoo! \footnote{\url{http://webscope.sandbox.yahoo.com}}. For the first time in this paper, this corpus is used to train skip-gram models. In contrast to the previous corpora, the description of real-world images contains spontaneous word usage which is directly related to visual content. Hence, we expect it to provide embeddings able to capture visual properties. 
\vspace{-3mm} 
\end{enumerate}
\begin{table}[ht]
  \centering
  \footnotesize
  \begin{tabular}{| l | l  | l | l| l |}
  \hline
  \textbf{Language} & \textbf{Abbrev.} & \# \textbf{Concepts} & \# \textbf{Pairs} & \# \textbf{Images} \\ \hline   
  {English} & EN & 4,421 &  1,109,467 & 447,997 \\ 
  {Spanish} & ES & 3,381 & 97,862 & 37,528 \\ 
  {Italian} & IT & 3,349 & 44,794 & 25,664\\ 
  {French} &  FR & 2,349 & 34,747 & 16,807\\ 
  {Chinese} & ZH & 504   & 21,049 & 5,562\\ 
  {German} &  DE & 804   & 14,635 & 7,335\\ 
  {Dutch}   & NL & 348   & 3,491 & 2,226\\ 
  {Russian} & RU & 129   & 1,536 & 800\\ 
  {Turkish} & TR & 231   & 941 & 638\\ 
  {Polish}  & PL & 63    & 727 & 477 \\
  {Persian} & FA & 15    & 56 & 34\\ 
  {Arabic} &  AR & 29    & 46 & 23\\ 
  \hline
  \end{tabular}
  \vspace{-2mm}
  \caption{ANP co-occurrence statistics for 12 languages, namely the number of concept tags and number of images with concept tags.}
  \label{gt_stats}
  \vspace{-5mm}
\end{table}

For the Google News corpus, we used pre-trained embeddings of 300 dimensions with a context window of 5 words provided by \cite{mikolov13}. For the other corpora, we trained the skip-gram model with a context window $w$ of 5 and 10 words, fixing the dimensionality of the word embeddings to 300 dimensions. In addition to training the vanilla skip-gram model on word tokens we propose to train each of the corpora (except Google news due to lack of access to original documents used for training) by treating ANPs (concepts) as unique tokens. This pre-processing step allows the skip-gram model to directly learn  concepts embeddings while taking advantage from the word contextual information over the above corpora. 

\subsection{Distributed Concept Representations}
\label{concept_emb}
To represent concepts in a semantic space we use the  word embeddings in the pivot language (English), and compose the representation of a concept based 
on its compound words. Each sentiment-biased visual concept $c_{i}$ comprises zero or more adjective and one or more noun words (as translation does not necessarily preserve the adjective-noun pair structure of the original phrase).
 Given the word vector embeddings, $\vec x_{\text{adj}}$ and $\vec x_{\text{noun}}$, we compute the concept embedding $\vec c_{i}$ using the sum operation for composition: $\vec c_{i} = \vec x_{\text{adj}} + \vec x_{\text{noun}}$ or the concept embedding $\vec c_{i}$ which is directly learned from the skip-gram model. This enables the comparison of multilingual concepts using the pivot language (English). At this stage, we note that there are several
other ways to define composition of short phrases, e.g.~\cite{baroni10,socher12,mikolov13}; however, here, we focus on the type of corpora used for obtaining the word embeddings rather than on the composition function.

\begin{table*}[htp]
  \footnotesize
  \centering
  \begin{tabular}{| l | r |  r | r | r | r | r | r | r |r |r | r | r |}
  \hline
  \textbf{Method} \textbackslash \ \textbf{Language} & \textbf{EN} & \textbf{ES} & \textbf{IT} & \textbf{FR} & \textbf{ZH} & \textbf{DE} & \textbf{NL} & \textbf{RU} & \textbf{TR}  & \textbf{PL} & \textbf{FA} & \textbf{AR} \\ \hline     
  wiki ($w$=10)          & 3.81   & 5.62  & 6.47  &  7.18  & 5.30 & 8.33 & 11.65 & 14.67 & 19.59 & 16.62 & 17.25 & 31.17 \\
  wiki-anp ($w$=10)      & 3.46   & 5.38  & 6.33  &  7.20  & 4.98 & 8.56 & 11.99 & 15.26 & 20.97 & 17.14  & 19.31 & 35.15 \\
  wiki-anp-l ($w$=10)    & 3.27   & 4.78  & 6.49  &  7.29  & 4.57 & 8.57 & 13.54 & 16.05 & 24.30 & 22.05 & 21.47 & 38.40 \\ \hline

  wiki\_rw ($w$=10)       & 10.17  & 12.01 & 12.08 &  12.11 & 13.62& 12.98& 11.02 & 13.74 & \textbf{12.71} & \textbf{12.28} & \textbf{6.51}  & \textbf{16.16} \\  
  wiki\_rw-anp ($w$=10)   & 3.79   & 5.54  & 6.38  &  7.23  & 5.16 & 8.53 & 11.67 & 14.94  & 19.79 & 16.48 & 17.91 & 32.34 \\   
  wiki\_rw-anp-l ($w$=10) & 3.57   & 4.90  & 6.43  &  7.21  & 4.90 & 7.91 & 13.28 & 15.27  & 23.29 & 21.15 & 20.15 & 34.59 \\ \hline

  flickr ($w$=10)          & 6.27   & 6.75  & 7.23  &  7.84  & 6.91 & 9.03 & \textbf{10.31} & \textbf{13.59} & 15.83 & 13.41  & 10.36 & 24.98 \\ 
  flickr-anp ($w$=10)      & 3.38   & 4.81  & 6.89  &  6.59  & 4.69 & \textbf{7.85} & 11.33 & 14.05  & 18.66 & 16.26 & 15.61 & 31.43\\  
  flickr-anp-l ($w$=10)    & 2.72   & 4.12  & 5.95  &  6.73  & 4.04 & 8.55 & 14.09 & 14.59  & 25.00 & 22.23  & 21.12 & 34.92 \\ \hline 

  gnews ($w$=5)            & 4.59   & 5.81  & 6.85  &  7.51  & 5.63 & 8.76 & 11.08 & 14.02 & 18.29 & 14.88 & 14.08 & 28.61 \\   
  wiki ($w$=5)             & 3.01   & 5.08  & 6.16  &  7.04  & 4.83 & 8.30 & 12.34 & 15.07 & 21.16 & 17.57 & 19.30 & 35.43 \\
  wiki-anp ($w$=5)         & 2.91   & 5.01  & 6.09  &  7.10  & 4.71 & 8.36 & 12.39 & 15.53 & 21.91 & 17.79 & 20.86 & 37.42 \\
  wiki-anp-l ($w$=5)       & 2.73   & 4.56  & 6.36  &  7.23  & 4.30 & 8.42 & 13.71 & 16.33 & 25.06 & 22.66 & 22.40 & 40.53 \\ \hline

  wiki\_rw ($w$=5)         & 5.70   & 7.36  & 8.12  &  8.47  & 8.51 & 9.48 & 10.52 & 13.60 & 15.34 & 13.43 & 10.12 & 22.40 \\   
  wiki\_rw-anp ($w$=5)     & 3.20   & 4.99  & 6.08  &  7.04  & 4.65 & 8.32 & 12.22 & 15.21 & 21.37 & 17.49 & 19.26 & 36.30 \\   
  wiki\_rw-anp-l ($w$=5)   & 3.03   & 4.58  & 6.35  &  7.21  & 4.55 & 8.47 & 13.74 & 15.78 & 24.50 & 22.18 & 21.24 & 37.86 \\ \hline

  flickr ($w$=5)           & 5.48   & 6.19  & 6.79  &  7.53  & 6.19 & 8.79 & 10.64 & 13.71 & 16.60 & 14.03 & 11.87 & 28.04 \\
  flickr-anp ($w$=5)       & 2.87   & 4.52& \textbf{5.85}  &  \textbf{6.56}  & 4.41 & 7.91 & 11.85 & 14.34 & 20.18 & 17.14 & 16.67 & 34.31 \\ 
  flickr-anp-l ($w$=5)     & \textbf{2.21} & \textbf{4.12}& 6.04  &  6.84  & \textbf{3.94} & 8.28 & 14.66 & 15.54  & 26.10 & 23.16 & 21.82 & 36.85 \\ 
  
   \hline
  \end{tabular}
  \vspace{-2mm}
  \caption{Comparison of the various concept embeddings on visual semantic relatedness in terms of MSE (\%) per language (lower is better). The embeddings are from Flickr (`flickr'), Wikipedia (`wiki') and Wikipedia + Reuters + Wall Street Journal (`wiki-rw') trained on a context window of $w \in \{10,5\}$ words using words as tokens or words and ANPs as tokens (`-anp'). All embeddings use the sum operation of noun and adjective vectors to compose ANP embedding for a given ANP except from the ones abbreviated with `-l' which use instead learned ANP embeddings when available.  The best score per language is marked in bold.}
  \label{sem_rel_lang}
  \vspace{-4mm}
\end{table*}

\section{Evaluation}
\subsection{Visual Semantic Relatedness} \label{gt}
\begin{figure}[ht]
\vspace{-1mm}
\centering
\hspace{-2mm} \includegraphics[scale=0.61]{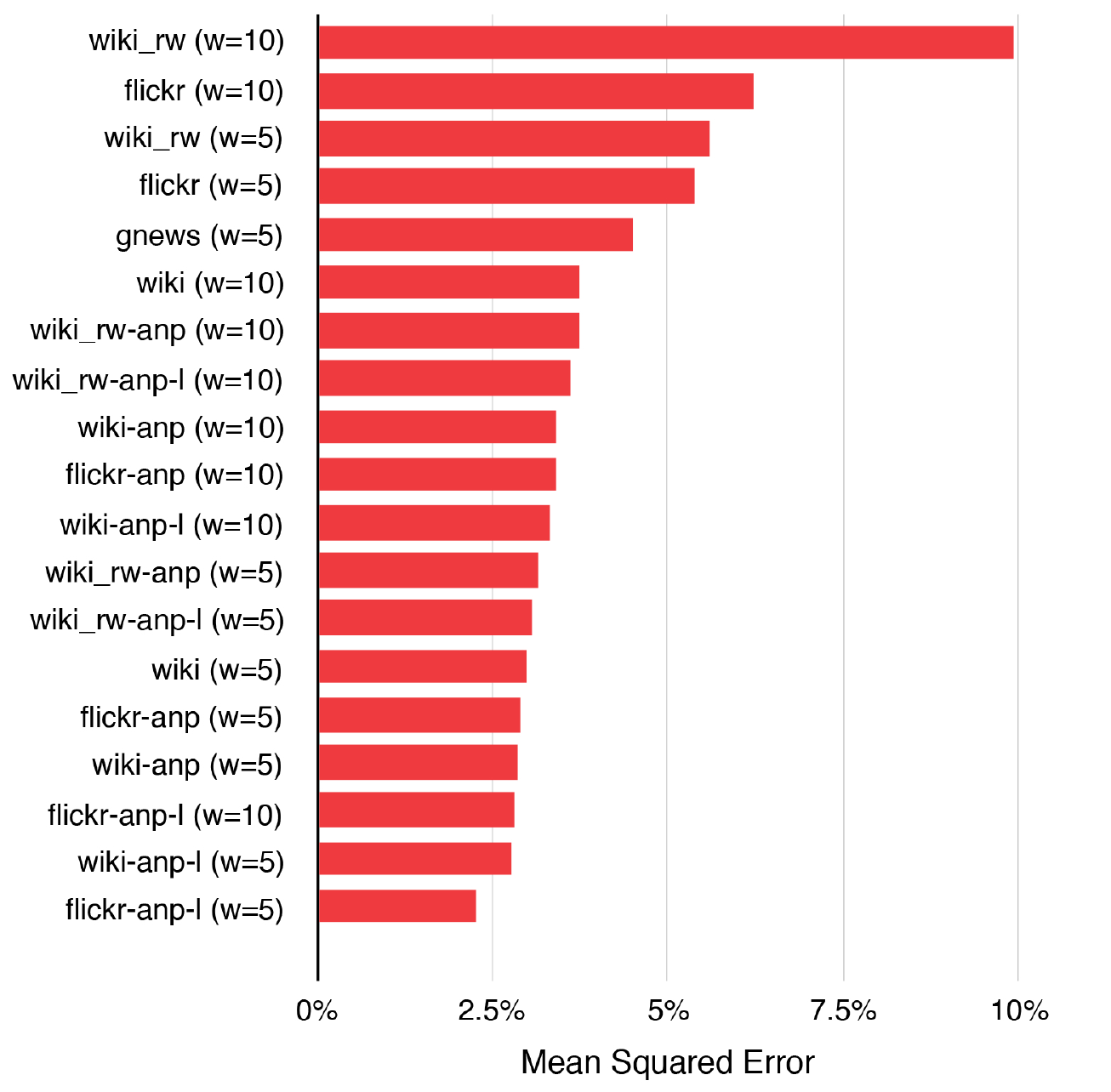}
\vspace{-3mm} 
\caption{Comparison of the various concept embeddings over all languages on visual semantic relatedness in terms of descending MSE (\%). For the naming conventions please refer to Table \ref{sem_rel_lang}. }
\vspace{-6mm}
\label{sem_rel}
\end{figure}

Evaluating word embeddings learned from text is typically performed on tasks such as semantic relatedness, syntactic relations and analogy relations \cite{word2vec}. These tasks are not able to capture concept properties related to visual content. For instance, while \emph{deserted beach} and \emph{lonely person} seem unrelated according to text, in the context of an image they share visual semantics. An individual person in a deserted beach gives to a remote observer the impression of loneliness. To obtain groundtruth for visual semantic distance  we collected co-occurrence statistics of ANPs (concepts) translated in English from 12 languages by analyzing the MVSO image tags (1,000 samples per concept), as displayed in Table \ref{gt_stats}. We compute the visually anchored semantic distance using cosine distance between two co-occurrence vectors: $d(\vec h_{i}, \vec h_{j}) = 1 - \text{cosine}(\vec h_{i},\vec h_{j})$.  We now compare the performance of the various concept embeddings of Section \ref{sec:embed} on the visual semantic relatedness task. Figure \ref{sem_rel} displays their performance over all languages in terms of Mean Squared Error (MSE), and Table \ref{sem_rel_lang} displays their performance per language accoring to the MSE score for all the pairs of concept embeddings $\vec c_{i}$ and $\vec c_{j}$, as:
\vspace{-1mm}
\begin{equation}
 \text{MSE} = \frac{1}{T} \sum^N_{i} \sum^{|\{i,\ldots, N\}|}_{j:j\neq i\ \&\ Uij\neq 0} (d(\vec c_{i}, \vec c_{j}) - d(\vec h_{i}, \vec h_{j})^2,
 \label{sdist}
\end{equation}
\noindent
where $U_{ij}$ is the co-occurrence between concepts $i$ and $j$, and $T$ is the total number of comparisons, that is  $T=\frac{(N^2 - N - |\{Uij = 0\}|)}{2}$. Note that to avoid redundancy, we have excluded from the comparisons the diagonal and the lower triangular matrix of the concept co-occurrence matrix  ($N \times N$). This does not affect the average mean squared error calculation because the cosine distance is symmetric. From the comparisons $T$ we further exclude ANP pairs with zero co-occurrence to penalize embeddings which tend to separate most of the concepts apart.  

The best performance over all languages (Figure \ref{sem_rel}) is achieved by the flickr-anp-l ($w$=5) embeddings, followed by the wiki-anp-l ($w$=5)  embeddings. The superior performance of flickr-anp-l ($w$=5) is attributed to its ability to learn directly the embedding of a given concept (ANP). The lowest performance is observed by wiki-reu-wsj ($w$=10) and flickr ($w$=10). The larger context ($w$=10) performed worse than the smaller context ($w$=5); it appears that the semantic relatedness prediction over all languages does not benefit from large contexts. When the concept embeddings are evaluated per language in Table \ref{sem_rel_lang} we obtain slightly different ranking of the methods. In the languages with the most data, namely English (EN), Spanish (ES), Italian (IT), French (FR) and Chinese (ZH), the ranking is similar as before, with flickr-anp-l ($w$=5), flickr-anp ($w$=5) and wiki-anp ($w$=5), wiki-anp-l ($w$=5) embeddings having the lowest error in predicting semantic relatedness. 

Generally, we observed that for well-resourced languages the quality of concept embeddings learned by a skip-gram model improves when the model is trained using ANPs as tokens (both when using directly learned concept embeddings or composition of word embeddings with sum operation). Furthermore, the usage of learned embeddings abbreviated with $-l$ on the top-5 languages outperforms on average all other embeddings in English, Spanish and Chinese languages and performs similar to the best embeddings on Italian and French. In the  low resourced languages the results are the following: in German (DE) language the lowest error is from flickr-anp ($w$=10), in the Dutch (NL) and Russian (RU) is the flickr ($w$=10).  Lastly, the lowest error in the Turkish (TR), Persian (FA) and Arabic (AR) languages is from wiki-reu-wsj ($w$=10). It appears that for the languages with small data the large context benefits the visual semantic relatedness task. 

Moreover, the performance of embeddings with a small context window ($w=5$), are outperformed by the ones that use a larger one ($w=10$) as the pool of examples of the languages narrows. This is likely due to the different properties which are captured by different context windows, namely more abstract semantic and syntactic relations with a larger context window and more specific relations with a smaller one. Note that the co-occurrence of concepts in MVSO images is computed on the English translations and hence some of the syntactic properties and specific meaning of words of low-resourced languages might have vanished due to errors in the translation process. Lastly, the superior performance of the embeddings learned from the Flickr 100M corpus in the top-5 most resourced languages, validates our hypothesis that word usage directly related to the visual content helps to learn concept embeddings with superior visual semantic properties.

\subsection{Multilingual Concept Clustering}
Given a common way to represent multilingual concepts, we are now able to cluster them.
We normalize the concept vectors to perform $k$-means clustering over Euclidean distances. We perform part-of-speech tagging on the translation to extract the representative noun or adjective with TreeTagger \cite{schmid_1994}.  We then experimented with two types of clustering approaches: a one-stage and a two-stage approach. The \textbf{one-stage} approach directly clusters all the concept vectors using $k$-means. The \textbf{two-stage} clustering operates first on the noun or adjective word vectors and then second on concept vectors. Here, we first split the translated concepts into different groups based on the clusters their nouns belong to, and then run another round of $k$-means clustering within the groups formed in the first stage. In the case when a translation phrase has more than one noun, we select the last noun as the representative and use it in the first stage of clustering. The second stage uses the sum of vectors for all the concepts that comprise that translation for clustering. We also experimented with first clustering by adjectives and then by nouns using the same process. 

We experimented with different number of clusters $k$. For the two-stage clustering, we adjust the number of clusters $k$ in the last stage based on the number of concepts in each first-stage cluster, e.g.~concepts in each noun-cluster ranged from 3 to 253 in one setup, to bring the final number of clusters in two-stage clustering to our desired number for fair comparison of clustering with different setups. We ended up with clusters such as \emph{beautiful music}, \emph{beautiful concert}, \emph{beautiful singer} that clusters concepts like \emph{musique magnifique} (French), \emph{bella musica} (Italian), \emph{ bellissimo concerto}. While noun-first clustering brings concepts that talk about similar objects, e.g.~estate/unit/property/building, adjective-based clustering yields concepts about similar and closely related emotions, e.g.~grateful/festive/joyous/floral/glowing/delightful (examples are from two-stage clustering with the Google News corpus).

To evaluate the clustering of affective visual concepts, we consider two dimensions: (1) \emph{Semantics}: ANPs are concepts, so we seek a clustering method to group ANPs with similar semantic meaning, such as for example \emph{beautiful woman} and \emph{beautiful lady}, (2) \emph{Sentiment}: Given that ANPs have an \emph{affective} bias, we need a clustering method that groups ANPs with similar sentiment values, thus ensuring the integrity of ANPs sentiment information after clustering. To evaluate clustering over both dimensions, we score clustering methods using two evaluation metrics specifically designed for this task, one for semantics and one for sentiment, and then combine the resulting ratings into a single score.
\begin{table}[htp]
\footnotesize
\centering
\begin{tabular}{| l | l  | l | l | l |}
\hline
   \textbf{Method} & \textbf{Embeddings} &  {$sen_C$} & {$sem_{C}$} & $\mu$ \\\hline

2-stage\_noun     & wiki (w=10)     & 0.511 & 0.588 & 0.549\\
2-stage\_noun     & wiki\_rw (w=10) & 0.529 & 0.604 & 0.566\\
2-stage\_noun     & flickr  (w=10)  & 0.538 & 0.528 & 0.533 \\ \hline

2-stage\_noun     & wiki (w=5)      & 0.534 & 0.586 & 0.560\\ 
2-stage\_noun     & wiki\_rw (w=5)  & 0.510 & 0.614 & 0.562\\
2-stage\_noun     & flickr  (w=5)   & 0.526 & 0.513 & 0.519\\ 
2-stage\_noun     & gnews (w=5)     & 0.309 & 0.569 & 0.439\\ \hline

2-stage\_adj      & wiki (w=10)     & 0.483 & 0.567 & 0.524\\
2-stage\_adj      & wiki\_rw (w=10) & 0.476 & 0.536 & 0.506\\
2-stage\_adj      & flickr  (w=10)  & 0.459 & 0.536 & 0.497\\ \hline 

2-stage\_adj      & wiki (w=5)      & 0.581 & 0.930 & 0.755\\
2-stage\_adj      & wiki\_rw (w=5)  & 0.472 & 0.560 & 0.516 \\
2-stage\_adj      & flickr  (w=5)   & 0.455 & 0.519 & 0.487 \\ 
2-stage\_adj      & gnews (w=5)     & 0.178 & 0.522 & 0.350\\ \hline

1-stage           & wiki-anp (w=10)     & 0.240 & 0.576 & 0.408 \\
1-stage           & wiki\_rw-anp (w=10) & 0.257 & 0.508 & 0.382\\
1-stage           & flickr-anp (w=10)   & \textbf{0.262} & \textbf{0.489} & 0.375 \\ \hline

1-stage           & wiki-anp (w=5)      & 0.250 & 0.583 & 0.416\\
1-stage           & wiki\_rw-anp (w=5)  & 0.281 & 0.522 & 0.402\\
1-stage           & flickr-anp (w=5)    & 0.280 & 0.502 & \textbf{0.391}\\ \hline

\end{tabular}
\vspace{-2mm}
\caption{Sentiment and semantic consistency of the clusters from the two-stage and one-stage $k$-Means clustering methods with $k=1000$, trained with the various concept embeddings. }
\vspace{-3mm}
\label{coherence_results}
\end{table}

\noindent \textbf{Semantic Consistency}. Each clustering method produces $k$ ANPs clusters, out of which $C$ contains two or more ANPs. For each of these multi-ANP clusters, each with $N_c$ ANPs, we compute the average visually grounded semantic distance (Eq.~\ref{sdist}) between all pairs of ANPs, and then average it over all $C$ clusters, thus obtaining a Semantic Consistency $\text{sem}_{C}$ metric for a given clustering method:
\vspace{-1mm}
\begin{equation}
  \text{sem}_{C}=\frac{1}{C}\sum_{c=1}^C \frac{ \sum^{|\{i,\ldots, N_c\}|}_{j:j\neq i\ \&\ Uij\neq 0} d(\text{ANP}_{c,i}, \text{ANP}_{c,j}) }{N_c}
\end{equation}
\\\textbf{Sentiment Consistency}. For each multi-ANP cluster $c$, we compute a \emph{sentiment quantization error}, namely the average difference between the sentiment of each ANP in the cluster, and the average sentiment of the cluster. So, 
given $\text{sen}_c= {\sum_{i=1}^{N_c} \text{sen}(\text{ANP}_i)}/{N_c}$ as the average sentiment for a cluster$c$, 
we obtain a Sentiment Consistency $sen_{C}$ metric for a given clustering method as follows:
\vspace{-1mm}
\begin{equation}
  \text{sen}_{C}=\frac{1}{C}\sum_{c=1}^C \frac{ \sum_{i=1}^{N_c} (\text{sen}(\text{ANP}_{c,i})-\text{sen}_c)^2 }{N_c}
\end{equation}
\textbf{Clustering Evaluation Results}.
We evaluate all the clustering methods using these two scoring methods and an overall \emph{consistency} metric (average of semantic and sentiment consistencies). The lower the value of the metrics, the higher the quality of the clustering method. We observe that semantic consistency and sentiment consistency are actually highly related. When we correlate the vector containing semantic consistency scores for all clustering methods with the vector containing sentiment consistency scores, we find that the Pearson's coefficient is around 0.7, thus suggesting that the higher the semantic relatedness of the clusters resulting from one method, the higher their respective sentiment coherence.

We observed that as the total cluster size $k$ increases, the consistency within a cluster also increases, even though the images and ANPs come from different languages and though the source of training corpus for word vectors did not always come from social content. Word vectors trained on Flickr data provided clustering with top consistency in all evaluation measures. Based on the analysis of sentiment consistency, we observed that 9 out of top 10 two-stage clustering setups were based on adjective-first clustering (only 9th being noun-first Google News clustering with 1000 clusters). This confirms our intuition that similar sentiments are clustered together when we first cluster ANPs with similar adjectives. Lastly, the vectors trained on full ANP tokens lead to increased semantic consistency, similar to on the full corpus (Section \ref{gt}).

\section{Portrait Concept Clustering}
The proposed multilingual concept clustering framework can be a useful tool for exploring and analyzing any large, multilingual collections of visual concepts. As an example application, we applied this concept matching framework to study  how affective concepts attach to human \emph{portraits}, i.e.~photos with faces, through the viewing lens of different languages. 

\subsection{Portrait-based Sentiment Ontology}
Portrait and face-centric photography has been a subject of research in multiple disciplines for years. Facial perception is among the most developed human capabilities, where our brains even contain a dedicated sub-network of neurons for face processing \cite{freiwald2014neurons}. Recently, computational understanding portrait modeling has attracted much attention from the multimedia community, e.g.~in computational aesthetics \cite{redi2015thebeauty}, animated GIFs \cite{jou2014gifs}, and social dynamics \cite{bakhshi14faces}.
Here, we seek to unpack what sentiment-biased visual concepts, specifically ANPs, languages attach to faces.

\vspace{3pt}
\noindent \textbf{Face Detection and ANP Filtering.}
To obtain a corpus of visual concepts relating to faces, we ran a frontal face detector \cite{liao_2015} which projects images onto a normalized pixel difference feature space and performs quadtree-based face detection.
A total of 3,858,869 faces were detected across the 7,368,364 images in the MVSO image dataset \cite{Jou2015}. Over 53.67\% of these detections came from the English image subset (2,071,078 detections), where the next leading language subset was Spanish at 23.68\% (913,596 detections). We then computed a \emph{portrait score} for each ANP which we define as the ratio of detected faces to all images in each ANP. We then selected the subset of ANPs whose portrait score was greater than 0.6. To ensure statistical significance, we only considered languages with 20 or more face-dominated ANPs: Turkish, Russian, German, Chinese, French, Spanish, Italian and English. Of the 11,832 concepts from the full MVSO dataset, we retained 2,345 face ANPs. We found that in general, detected faces from the French and German datasets we larger in size on average than other languages. In addition, images originating from the Italian subset typically contained more than one person while images in the Chinese and Turkish subset tended to contain mostly single-subject portraits.

\vspace{3pt}
\noindent \textbf{Concept Sentiment and Face Portraits.}
To explore the sentiment correlations of different languages to the presence of faces, we computed the Pearson's correlation coefficient $\rho$ for each language between ANP portrait scores and the ANP sentiment values, as shown in Table \ref{tab:sentstatistics}. 
The higher the correlation, the higher the tendency of a given language to associate positive sentiment with a face image.
Here, for all languages except Turkish, the presence of portraits in an ANP image pool tended to be positively correlated with the ANP sentiment.
In particular, the languages having the strongest tendency to attach positive sentiments to portraits are Russian and Chinese.

\subsection{Multilingual Portraits}
We sought to investigate how similar/different languages are with regard to affective visual concepts (ANPs) and their face images.
We clustered face ANPs from the subset of eight languages using our approximate match-based clustering techniques, and evaluated different clustering approaches to find that the \emph{single-step} clustering over Flickr-trained word2vec vector with $w=5$ and $k=1000$ gave us the best results when combining semantic and sentiment consistency metrics.
This method output 1,000 multilingual clusters of affective visual concepts related to portraits.
This provides a powerful tool to analyze the visual concept preferences for different languages, i.e.~if the ANPs of two languages fall often in similar clusters, such languages tend to attach similar concepts to face images.
\begin{table}[ht]
  \centering
  \resizebox{\linewidth}{!}{
  \begin{tabular}{|l|c|c|c|c|c|c|}
  \hline
   & \textbf{$\rho$(face,sent)} & \textbf{sent(faces)} & \textbf{sent(all)} & \textbf{diff(\%)} & \textbf{face size(\%)} & \textbf{\#faces(\%)} \\ \hline
   Turkish & 0.00         & 3.54       & 3.55 & -0.26   & 63.46       & 0.95 \\
   Russian & \textbf{0.23}        & \textbf{4.13}      & 3.67 & \textbf{12.48}   & 58.25       & 1.23 \\
   German  & 0.18         & 3.75       & 3.39 & 10.70   & 65.49       & 0.99 \\
   Chinese & \textbf{0.23}        & \textbf{4.30}      & 3.57 & \textbf{20.33}   & 64.12       & 0.93 \\
   French  & 0.14         & 3.48       & 3.32 & 4.79    & \textbf{65.88}       & 1.01 \\
   Spanish & 0.16         & 3.72       & 3.44 & 7.93    & 65.08       & 1.23 \\
   Italian & 0.15         & 3.75       & 3.38 & 10.97   & 61.72       & \textbf{1.37} \\
   English & 0.15         & 3.51       & 3.32 & 5.49    & 56.78       & 1.04 \\ \hline
  \end{tabular}}
  \vspace{-2mm}
  \caption{Sentiment statistics per language (face and all ANPs).}
  \label{tab:sentstatistics}
  \vspace{-1mm}
\end{table}

\vspace{3pt}
\noindent \textbf{Languages, Sentiments, Face Sizes.}
Among the clusters found by this method, around 60\% are monolingual clusters: English and Spanish are the languages with the highest percentage of monolingual clusters ($\sim$32\% and 31\% respectively), probably due to the large number of ANPs, implying a wider vocabulary compared to other languages. Around 22\% of the face ANP clusters are bilingual, out of which 40\% contains 2 of the 4 Western-most languages  in the corpus (French, Italian, English and Spanish), while the others contain a mixture of other languages. The remaining 18\% of clusters show three languages or more. But what is the relation between multilingual clusters and portrait sentiment? Do languages agree more on positive or negative sentiment for similar visual concepts? To answer these, we computed the correlation between the number of languages falling into each cluster and the average sentiment of the ANPs in that cluster. These two dimensions statistically significantly correlate with a coefficient of 0.13, showing that different languages tend to associate similar visual concepts to portraits when concepts carry positive sentiment.

We also analyze the relation between cluster multilingual-ness and face sizes.
Do visual concepts shared by different languages refer to portraits with bigger or smaller faces? We compute here the correlation between the number of languages falling into each cluster and the average face sizes for the ANPs in that cluster. The correlation coefficient stands at 0.17, showing that, the bigger the average face sizes of portraits related to a visual concept, the higher the possibility that different languages share such concept.

\begin{figure}[t]
  \centering
  \includegraphics[width=0.75\linewidth]{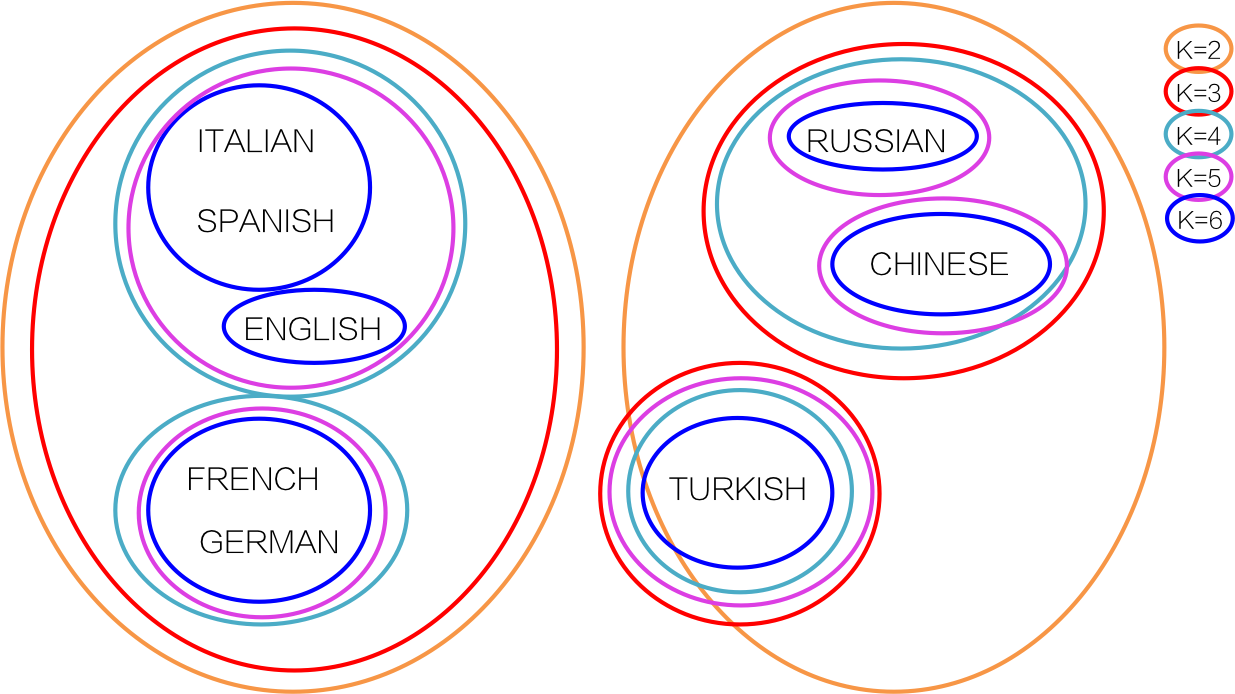}
  \vspace{-2mm}
  \caption{Groups of languages according to their similarity when associating affective concepts to faces. Each color corresponds to the output of a different clustering granularity $k$.}
  \vspace{-4mm}
  \label{fig:languageclusters}
\end{figure}

\vspace{3pt}
\noindent \textbf{Clustering Analysis.}
Which languages are more similar when tagging portraits?
To further understand language-specific concepts used when tagging face images, we perform a multivariate analysis of language distribution across visual concepts. To better understand which groups of languages tend to attach similar affective concepts to face images, we proceed as follows. We create eight $k$-dimensional vectors, one for each language. Each element of such vectors corresponds to the number of ANPs falling into each cluster, normalized by the total number of ANPs for a given language. Finally, we cluster these vectors using $k$-means with cosine distance, progressively raising $k$ from 2 to 6, thus separating languages into different groups, as shown in Figure \ref{fig:languageclusters}. The binary subdivision of the languages in two clusters ($k=2$) shows immediately a clear separation between more Eastern (Turkish, Chinese, Russian) versus Western (Italian, Spanish, French, English, German) languages. When we raise the number of clusters to 3, Turkish is the first one to detach from the Eastern clusters, suggesting that Turkish images tend to have more unique ways to assign concepts to portraits. Within the Western languages, when $k=4$ we see a separation: Italian gets clustered with Spanish and English, while French gets clustered with German. Chinese and Russian become  independent clusters for $k=5$. Finally, when raising $k$ to 6, Western languages split again: English becomes an independent cluster, leaving two bilingual clusters: Italian/Spanish and French/German.

\vspace{3pt}
\noindent \textbf{What Do Different Languages Say About Portraits?}
According to our clustering method, $\sim$3\% of the clusters contain five or more languages.
The limited size of this data allows us to proceed with manual inspection, to understand the topics of the ANPs falling in highly multilingual clusters.
The most highly multilingual cluster contains 8 languages and 20 ANPs: its main topic is about \emph{little guy} or \emph{little girl} (e.g.~\emph{piccola bimba} in Italian, or \emph{petit fille} in French). One interesting observation is that not all languages agree on the sentiment value for this concept: while Chinese and Turkish give a score slightly below 3, Italian, Spanish and Russian languages consider it a very positive concept, having respectively an average sentiment value of 4.0, 4.0 and 4.6, respectively, for this cluster. 
The second biggest cluster, spanning seven languages (all apart from Chinese) with 24 ANPs, contained concepts like \emph{gorgeous girl} in English or \emph{belle fille} in French.
Here, all languages agree on the highly positive value of this concept sentiment.
Other highly multilingual noun clusters contain concepts related to \emph{happy children}, \emph{young women}, \emph{healthy food}, \emph{beautiful women}, etc, and also negative concepts such as \emph{sexual violence}.

\section{Conclusions \& Future Work}
In this study, we showed that visual sentiment concepts from multiple languages can be effectively represented in a common semantic space using a pivot language (English) and existing advances in distributional semantics of words. The best results on the prediction of visually grounded semantic distance was achieved by a skip-gram model trained on real-world image metadata, namely titles and descriptions from the Flickr 100M corpus, with a summed combination of concept word embeddings or also by direct learning. This enabled multilingual clustering of visual sentiment concepts in 11 languages, and allowed us to better hierarchically organize an ontology in \cite{Jou2015} as well as provide deep analysis into portrait imagery from a multilingual perspective.

In the future, we plan to learn visually grounded concept embeddings in multiple languages directly from the groundtruth co-occurrence data (Section \ref{concept_emb}), e.g.~by modeling the proposed visual semantic distance. An essential element for achieving this goal is to semantically align multiple languages using parallel corpora, as it has been effectively done in previous studies in multilingual representation learning. Finally, we plan to evaluate such word and concept embeddings on visual sentiment concept prediction, concept recommendation and caption generation.

\scriptsize
\bibliographystyle{ieeetr}

\end{document}